**Multimodal Foundation Models For Echocardiogram Interpretation**

Matthew Christensen[1], Milos Vukadinovic[1,2], Neal Yuan[3], David Ouyang[1,4]

1. Department of Cardiology, Smidt Heart Institute, Cedars-Sinai Medical Center, Los Angeles, CA
2. Department of Bioengineering, University of California Los Angeles, Los Angeles, CA
3. Department of Medicine, University of California, San Francisco, CA; Division of Cardiology, San Francisco Veterans Affairs Medical Center, San Francisco, CA
4. Division of Artificial Intelligence in Medicine, Cedars-Sinai Medical Center, Los Angeles, CA

Correspondence: david.ouyang@cshs.org

**Abstract**

Multimodal deep learning foundation models can learn the relationship between images and text. In the context of medical imaging, mapping images to language concepts reflects the clinical task of diagnostic image interpretation, however current general-purpose foundation models do not perform well in this context because their training corpus have limited medical text and images. To address this challenge and account for the range of cardiac physiology, we leverage 1,032,975 cardiac ultrasound videos and corresponding expert interpretations to develop EchoCLIP, a multimodal foundation model for echocardiography. EchoCLIP displays strong zero-shot (not explicitly trained) performance in cardiac function assessment (external validation left ventricular ejection fraction mean absolute error (MAE) of 7.1%) and identification of implanted intracardiac devices (areas under the curve (AUC) between 0.84 and 0.98 for pacemakers and artificial heart valves). We also developed a long-context variant (EchoCLIP-R) with a custom echocardiography report text tokenizer which can accurately identify unique patients across multiple videos (AUC of 0.86), identify clinical changes such as orthotopic heart transplants (AUC of 0.79) or cardiac surgery (AUC 0.77), and enable robust image-to-text search (mean cross-modal retrieval rank in the top 1% of candidate text reports). These emergent capabilities can be used for preliminary assessment and summarization of echocardiographic findings.

**Introduction**

Echocardiography, or cardiac ultrasound, is the most common, non-invasive method of evaluating heart function and identifying heart disease. Echocardiography is routinely used to guide clinical decision-making,[1–3] and is crucial for disease diagnosis, risk stratification, and assessment of treatment response.[1,4] Recent work has used AI to improve the accuracy of echo measurements[5–7] and disease diagnoses,[8–10] however these approaches focus on individual tasks without the use of foundation models.[11]

Recent advances in AI have leveraged representation learning on large image and text corpuses to develop multimodal models that generalize beyond a narrow set of visual concepts.[12,13] These multimodal foundation models learn compact representations of images and text and can then be used to perform a wide variety of separate prediction tasks for which the model was never specifically trained ('zero-shot' tasks). Performance of foundation models on these indirect tasks are often more robust than with conventional convolutional neural networks.[14,15] In biomedical applications, foundation models have been developed to organize biological[16–18] and medical[19] concepts, including modality specific models for chest X-rays.[20,21] However, such model training has been bottlenecked by dataset size, and often limited to publicly available data that might not represent the range of disease severities and possible presentations.

In this work, we introduce EchoCLIP, a foundation model for echocardiography trained on a dataset of 1,032,975 echocardiogram videos across over a decade of clinical imaging. To assess the model's performance, we evaluated the model's 'zero-shot' ability to assess cardiac function, pulmonary artery pressure, and chamber size, as well as identify common intracardiac devices in both held-out internal test cohorts as well as independent external test cohorts. Cross-modal similarity comparison was used as a framework for assessing the model's ability to identify unique patients across time, identify clinically important changes as well as retrieve preliminary text interpretations of echocardiographic images.

**Results**

EchoCLIP is a medical language-image pretrained model finetuned with 1,032,975 unique video-text pairs from 224,685 echocardiography studies across 99,870 patients across a decade of clinical care (Table 1). To assess the importance of pretraining and architecture[22], different architectures and dataset configurations were compared (Supplementary Table 1). The final EchoCLIP model used a ConvNeXt image encoder and a BPE tokenizer. It has a context length of 77 tokens and was evaluated on a variety of zero-shot medical image interpretation tasks (Figure 1).

**Zero-shot medical image interpretation**

Without any supervised or task-specific training, we evaluated EchoCLIP's ability to perform medical image interpretation through zero-shot classification of videos in our internal held-out test set. EchoCLIP can identify intracardiac devices including percutaneous mitral valve repair with an AUC of 0.97 (95% CI: 0.97-0.98), transvenous aortic valve replacement (TAVR) with an AUC of 0.92 (95% CI: 0.91-0.92), and pacemaker/defibrillator leads with an AUC of 0.84 (95% CI: 0.84-0.85). Furthermore, EchoCLIP can zero-shot detect changes from normal cardiac chamber size, including severe dilation of the right ventricle with an AUC of 0.92 (95% CI: 0.91-0.92), right atrium with an AUC of 0.97 (95% CI: 0.97-0.98), left ventricle with an AUC of 0.92 (95% CI: 0.92-0.93), and left atrium with an AUC of 0.91 (95% CI: 0.90-0.92).

**External validation of zero-shot cardiac function and pressure assessment**

We further evaluated EchoCLIP's performance on zero-shot tasks including evaluation of cardiac function (left ventricular ejection fraction) and estimation of pulmonary artery pressure as reported by echocardiography. EchoCLIP predicts the left-ventricular ejection fraction on the held-out internal test dataset with an MAE of 8.4% and an MAE of 7.1% on an external test set of videos from the EchoNet-Dynamic dataset from Stanford Healthcare (Figure 2). Furthermore, EchoCLIP predicts estimated pulmonary arterial pressure with an MAE of 10.8 mmHg in the internal test dataset, and an MAE of 10.8 on the external test dataset (Figure 3).

**Mapping text to echocardiographic imaging**

A long context EchoCLIP-R model was optimized for retrieval using a custom cardiology-report-text-specific tokenizer (Table 2). Given an image from the held-out test cohort, EchoCLIP-R on average ranks the matching clinical report 209th out of 21,484 candidates. The correct report is present in the top 10 ranked reports 33.3% of the time. Going from text to image, the average rank of the matching video is

203 out of 21,484 and the correct video is present in the top 10 ranked videos 34.3% of the time. The ability to measure the similarity between pairs of echocardiograms can also be used to identify a unique patient across multiple studies (a difficult task for human clinicians) as well as identify clinical changes over time.

**Detection of clinically significant differences between videos**

Measuring the cosine similarity between EchoCLIP-R embeddings can be used to distinguish between image pairs from different patients (mean cosine similarity = 0.40, 95% CI: 0.40-0.40), images from the same patient across different timepoints (mean cosine similarity = 0.64, 95% CI: 0.64-0.65), and images from the same patient at the same time point (mean cosine similarity = 0.87, 95% CI: 0.86-0.87). This comparison results in an AUC of 0.86 (95% CI: 0.86-0.86) in identifying the same patients across different videos (Figure 5). Furthermore, the cosine similarity can also be used to distinguish when there was a significant clinical change, with echocardiogram studies occurring after cardiac surgeries and orthotopic heart transplants displaying a lowered similarity compared to studies occurring before such events. This dropoff in embedding similarity is significant enough that it can predict whether an echocardiogram occurs before or after cardiac surgery with an AUC of 0.77 (95% CI: 0.75-0.79), and before or after heart transplant with an AUC of 0.79 (95% CI: 0.76-0.82).

**Discussion**

Our results suggest that large datasets of medical imaging and expert adjudicated interpretations can serve as the basis for training medical foundation models. A single trained foundation model was able to successfully complete multiple zero-shot prediction tasks without task-specific training or fine-tuning, including left ventricular ejection fraction assessment and identification of intracardiac devices. Additionally, EchoCLIP-R displays emergent ability to perform tasks that human clinicians struggle with, such as identifying the same patient across different imaging studies. EchoCLIP-R is also capable of characterizing clinically significant changes over time, including heart transplants and other cardiac surgeries.

A key bottleneck in training medical foundation models has been limited data. Prior AI models were trained with a max of 150,000 echocardiogram videos,[23] while EchoCLIP is trained on over one million. By leveraging large historical clinical reporting databases, our approach minimizes tedious manual labeling and organization required for most supervised learning tasks. Additionally, by training EchoCLIP with data from one healthcare system and test its performance on an entirely separate external healthcare system, we were able to evaluate for generalizability for such foundation models. While task specific models still perform better on narrow tasks,[23] the performance of EchoCLIP on external validation data confirm its ability to assess cardiac function with similar accuracy to human clinicians.[24,25]

While not the first instance of a foundation model trained on healthcare or biomedical specific tasks,[26–28] this is the first such model specific for echocardiography, the most common modality for cardiovascular imaging. Future work will incorporate video encoders and leverage multiple views from the same echocardiographic study to provide more holistic evaluations of heart health. Enhancements such as upgrading EchoCLIP's visual encoder from an image-based model to a video-based model, adapting EchoCLIP for visual question answering, and implementation of automatic report generation are potential directions for future research.

Our results encourage further exploration of multimodal foundation modals for cardiology and medicine generally. Clinical datasets provide large corpuses of information about health, while different modalities provide adjunctive ancillary information that might improve upon understanding of cardiovascular health. Further efforts remain to leverage larger datasets and model architectures that can succinctly distill medical information.

# Methods

## Data Curation

The Cedars-Sinai Medical Center echocardiography laboratory performs clinical echocardiography for a wide range of indications ranging from asymptomatic pre-operative screening to evaluation for open heart surgery or heart transplant. A standard full resting echocardiogram study consists of a series of 50-150 videos and images visualizing the heart from different angles, locations, and image acquisition techniques (2D images, tissue Doppler images, and color Doppler images). Each echocardiogram study corresponds to a unique patient during a unique visit, however multiple similar videos might be obtained from the same view for the same patient during a single study. For EchoCLIP, we focused on the apical-4-chamber view (one of the most common and well-obtained ultrasound views) and organized a dataset of 1,032,975 unique video-caption pairs from 224,685 echocardiogram studies across 99,870 patients, collected between 2011 and 2022.

Data was split by patient across training, validation, and internal test datasets. We split the data into a training set containing 921,981 videos from 84,990 patients, a validation set containing 10,000 videos from 5,358 patients, and a test set containing 100,994 videos from 10,001 patients. A random subset (n = 5000) of the publicly released EchoNet-Dynamic dataset from Stanford Healthcare was used as an external test dataset. An automated preprocessing workflow was undertaken to remove extraneous text, ECG and respirometer information, and other information outside of the scanning sector and the input data was formatted in standardized 224x224 pixel RGB videos. This research was approved by the Cedars-Sinai Medical Center and Stanford Healthcare Institutional Review Boards.

## Model design and training

Model design and training was done in Python using the PyTorch deep learning library. Our training code is a fork of the OpenCLIP repository.[22] To find the best training configuration, we evaluated a variety of model architectures and training procedures (Supplementary Table 1), with the final model using the ConvNeXt architecture[29] for the image encoder and a decoder-only transformer for the text encoder. We initialize our model with weights pretrained on LAION-400M, and trained for 50 epochs, minimizing the CLIP loss. We warm up to an initial learning rate of 5e-5 over the course of 2,000 steps and then cosine decay to zero over the course of the training run. We use a batch size of 1,024 and train on a pair of Nvidia RTX A6000 48GB GPUs for approximately 2 weeks. Model checkpoints are saved every epoch, and at the end of training, the model checkpoint with the lowest validation mean cross-modal retrieval rank was used for testing. During training, a random frame is extracted from each video to be passed to the image encoder, with a different random frame from each video used per epoch as a form of data augmentation.

**Text tokenization**

A variety of text tokenization schemes were tested (Supplemental Table 1). EchoCLIP was trained using text tokenized by a Byte-Pair Encoding tokenizer[30] pretrained on the GPT2 data corpus, which encoded echocardiography reports with a mean of 530.3 (±154.7) tokens per report. We compared performance with a BPE tokenizer pretrained on the echocardiography reports. We noted that the echocardiography report text is often highly structured and repetitive, allowing tokenization with relatively few unique tokens representing long phrases. A custom-built echocardiography report tokenizer with more aggressive distillation of the data was designed.

Instead of searching for exact vocabulary matches in the report text, our template tokenizer uses regular expressions to allow nearly-similar lines of text to be efficiently encoded. For example, the text "Moderate left ventricular hypertrophy. Left ventricular ejection fraction is 60%" is converted into tokens referring to either cardiac structure or function (such as "<_ left ventricular hypertrophy>", "<left ventricular ejection fraction is _%>") as well as indicating severity ("mild", "moderate", or "severe") or quantity (60%, 2.5cm)By doing this, we were able to capture most of the variance present in our text reports with a vocabulary containing only 770 words and phrases, in addition to extra tokens for handling numbers and severity terms. After applying this custom tokenizer, the mean length of a tokenized report was brought down to just 63.8 (±26.7) tokens, approximately a nine-fold reduction compared to using CLIP's original BPE tokenizer.

**Retrieval and semantic search**

With EchoCLIP-R embeddings, we perform similarity search amongst other images and text to find similar examples based on cosine similarity with the query embedding. For semantic search, the image-to-text rank and text-to-image rank refers to the relative rank by cosine similarity of the true associated image or text among all same-modality candidates. We restricted our test dataset to one unique image-text pair for each report and then calculated the mean cross-modal rank. The mean cosine-similarity across the same patient in the same study, same patient across different studies, and different patients were evaluated to evaluate the model's ability to correctly identify the same patient and meaningful clinical changes. To evaluate a patient's trajectory before or after heart transplantation or cardiac surgery, we plotted the cosine similarity between echocardiograms acquired within 200 days of the procedure and the earliest echocardiogram within 200 days of the procedure. These similarity timelines were used to create Figure 5c and 5d. These similarity values can also be treated as zero-shot predictions of whether a given video was acquired before or after the procedure, and used to calculate an AUC score which quantifies EchoCLIP-R's ability to detect the effects of such procedures.

**Zero-shot evaluation**

We evaluate EchoCLIP on zero-shot tasks which are both binary classification tasks as well as continuous value regression tasks. For binary classification tasks, we constructed text prompts describing a positive case and compared the embedding of this prompt with the embeddings of videos in our test set. We treat the cosine similarity between the text embedding and the video embeddings as the model's zero-shot prediction. Ground-truth labels are extracted from the actual reports and used to calculate AUC and other performance metrics.

For regression tasks, we generated an ensemble of variations on the same text prompt only changing the relevant value in the text. For instance, "The left ventricular ejection fraction is estimated to be X%" or "LV ejection fraction is X%" text prompts were embedded for all integer values between 0 and 100. The cosine similarity between this prompt embedding and the image embedding was calculated, with the prediction being the median value of the top 20 percent of prompt embeddings. This is calculated for the first 10 frames of the video and the final predicted value is the average of the 10 per-frame predictions.

**Data availability**

The dataset of videos and reports used to train EchoCLIP is not publicly available due to its potentially identifiable nature. However, EchoNet-Dynamic, the dataset we used for external validation, is publicly available at https://echonet.github.io/dynamic/.

**Code availability**

Our model weights and code are available at https://github.com/echonet/echo_CLIP

# Tables

## Table 1

|  | Total | Training | Validation | Test |
|---|---|---|---|---|
| *n* | 224,685 | 195,082 | 8,119 | 21,484 |
| *Age (mean (SD))* | 66.26 (16.74) | 66.3 (16.7) | 65.8 (17.0) | 65.7 (16.9) |
| *Female = TRUE (%)* | 96,451 (42.9) | 83,700 (42.9) | 3,363 (41.4) | 9,388 (43.7) |
| *Race (%)* |  |  |  |  |
| *Native American* | 526 (0.2) | 456 (0.2) | 23 (0.3) | 47 (0.2) |
| *Asian* | 16,601 (7.5) | 14,450 (7.5) | 555 (6.9) | 1,596 (7.5) |
| *Black* | 29,546 (13.3) | 25,624 (13.3) | 1,104 (13.8) | 2,818 (13.3) |
| *Hispanic* | 22,424 (10.1) | 19,394 (10.0) | 842 (10.5) | 2,188 (10.3) |
| *Non-Hispanic White* | 133,399 (60.0) | 116,044 (60.1) | 4,699 (58.6) | 12,656 (59.6) |
| *Other* | 15,376 (6.9) | 13,243 (6.9) | 612 (7.6) | 1,521 (7.2) |
| *Pacific Islander* | 767 (0.3) | 688 (0.4) | 36 (0.4) | 43 (0.2) |
| *Unknown* | 3,700 (1.7) | 3,182 (1.6) | 149 (1.9) | 369 (1.7) |
| *AF* | 46,994 (20.9) | 41,214 (21.1) | 1,633 (20.1) | 4,147 (19.3) |
| *HF* | 75,358 (33.5) | 65,802 (33.7) | 2,764 (34.0) | 6,792 (31.6) |
| *HTN* | 90,738 (40.4) | 79,229 (40.6) | 3,250 (40.0) | 8,259 (38.4) |
| *CVA/TIA/TE* | 38,283 (17.0) | 33,475 (17.2) | 1,378 (17.0) | 3,430 (16.0) |
| *MI* | 14,983 (6.7) | 13,120 (6.7) | 514 (6.3) | 1,349 (6.3) |
| *CAD* | 55,659 (24.8) | 48,840 (25.0) | 2,040 (25.1) | 4,779 (22.2) |
| *PAD* | 23,369 (10.4) | 20,475 (10.5) | 838 (10.3) | 2,056 (9.6) |
| *DM* | 37,900 (16.9) | 33,226 (17.0) | 1,351 (16.6) | 3,323 (15.5) |
| *CKD* | 40,947 (18.2) | 35,960 (18.4) | 1,482 (18.3) | 3,505 (16.3) |
| *Prior Smoker* | 7,632 (3.4) | 6,593 (3.4) | 256 (3.2) | 783 (3.6) |

**Table 1.** Clinical characteristics of study cohort, reported by echocardiography study. AF = Atrial Fibrillation. HF = Heart Failure. HTN = Hypertension. CVA = Cerebrovascular Accident. TIA = Transient Ischemic Attack. TE = Thromboembolism. MI = Myocardial Infarction. CAD = Coronary Artery Disease. PAD = Pulmonary Artery Disease. DM = Diabetes Mellitus. CKD = Chronic Kidney Disease.

**Table 2**

| | Image Encoder | Tokenizer | MCMRR | LVEF, MAE | PAP, MAE | TAVR, AUC | MitraClip, AUC | Pacemaker, AUC |
|---|---|---|---|---|---|---|---|---|
| *CLIP* | ViT-B-32 | CLIP BPE | 10,743.0 | 20.8 (20.7-20.8) | 16.8 (16.8-16.9) | 0.46 (0.46-0.47) | 0.53 (0.52-0.54) | 0.51 (0.51-0.52) |
| *EchoCLIP* | ConvNeXt | CLIP BPE | 571.3 | **8.4 (8.3-8.4)** | **10.8 (10.8-10.9)** | **0.92 (0.91-0.92)** | **0.97 (0.97-0.97)** | **0.84 (0.84-0.84)** |
| *EchoCLIP-R* | ConvNeXt | Template tokenizer | **206.1** | 16.9 (16.8-17.0) | 17.5 (17.4-17.5) | 0.52 (0.51-0.52) | 0.81 (0.81-0.82) | 0.66 (0.65-0.66) |

**Table 2.** Main performance metrics. EchoCLIP achieves strong performance in several classification and regression tasks. EchoCLIP-R, while excelling at retrieval, does so at the cost of poorer performance on other zero-shot tasks. Retrieval ranks are out of 21,484 candidates. Ranges in parentheses indicate 95% confidence intervals bootstrapped with 1,000 random samples. MCMRR = Mean Cross-Modal Retrieval Rank. LVEF = Left Ventricular Ejection Fraction. PAP = Pulmonary Artery Pressure. TAVR = Transvenous Aortic Valve Replacement.

**Figures**

**Figure 1**

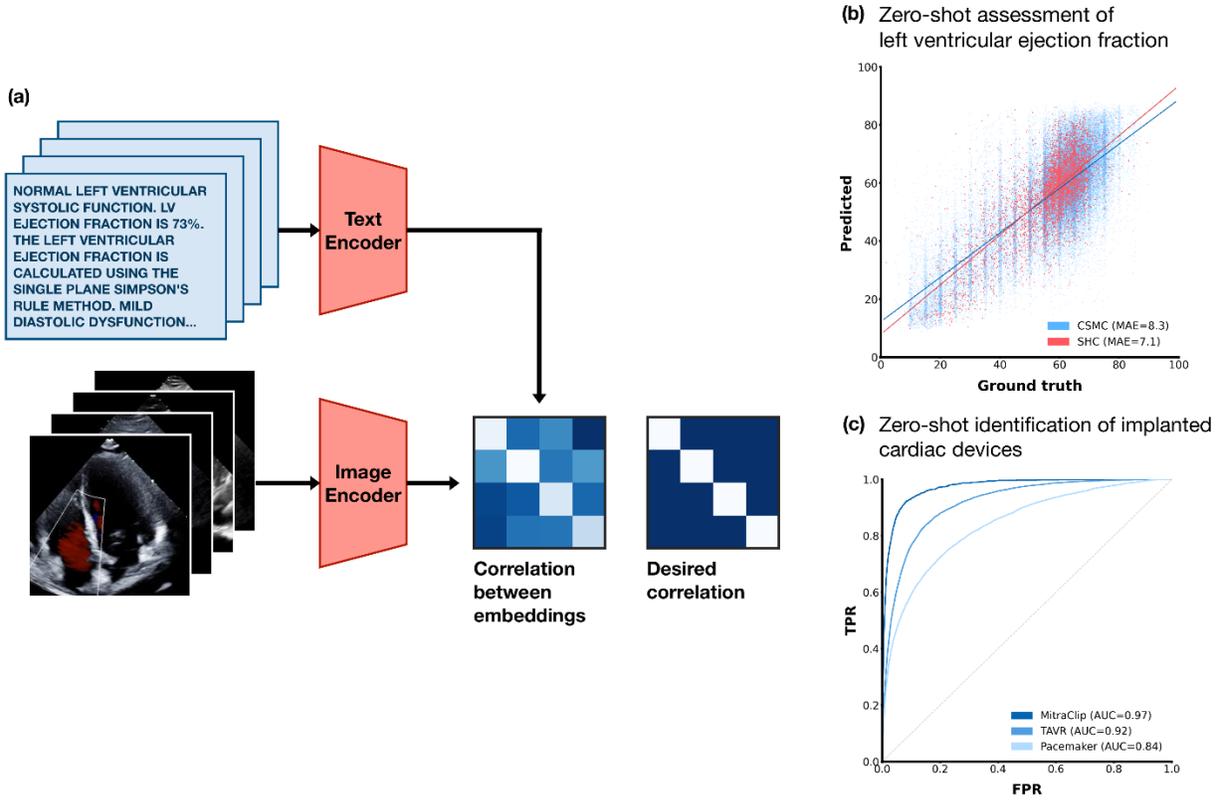

**Figure 1.** EchoCLIP workflow (a) EchoCLIP is a foundation model trained on one million echocardiogram videos across 11 years. It is composed of an image encoder for processing echocardiogram video frames and a text encoder for processing the corresponding physician interpretations. These two encoders project the images and interpretations into a joint embedding space. (b) The trained EchoCLIP model can zero-shot assess left ventricular systolic function in held-out test-set videos from Cedars-Sinai Medical Center (blue, n = 100,994) and Stanford Healthcare (red, n = 5,000). (c) Similarly, the trained EchoCLIP can identify implanted intracardiac devices including MitraClip, transvenous aortic valve replacement (TAVR) valves, and implanted pacemaker/defibrillator leads on held-out test videos from Cedars-Sinai Medical Center.

**Figure 2**

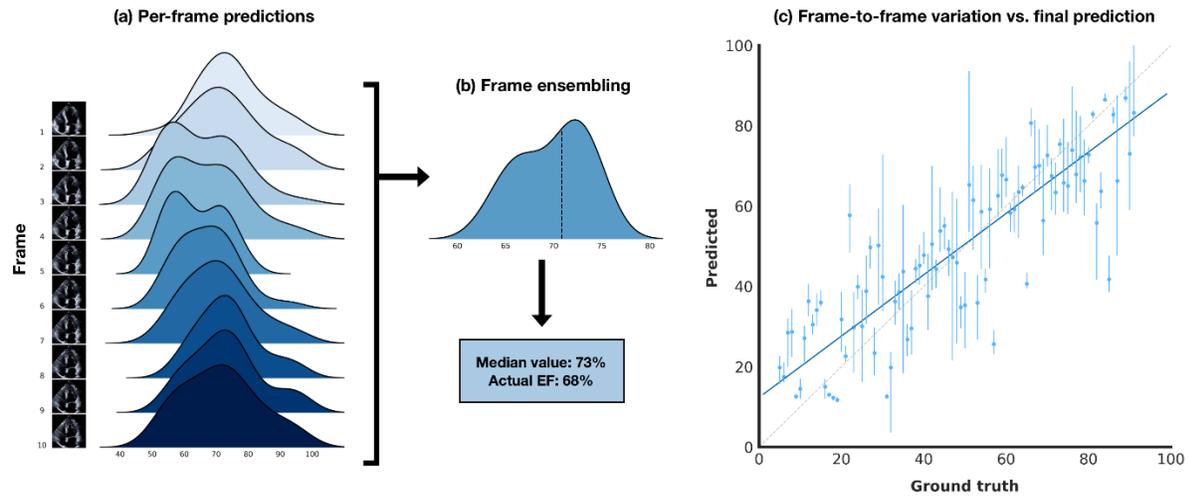

**Figure 2.** Frame level ensembling. (a) EchoCLIP zero-shot predicts continuous values from individual frames of an echocardiogram video. (b) The individual frame-level predictions are then ensembled for video-level prediction. (c) In the scatterplot, the points represent the final predicted values, while whiskers represent the full range of frame level predictions, demonstrating the importance of frame ensembling in improving accuracy on this task.

**Figure 3**

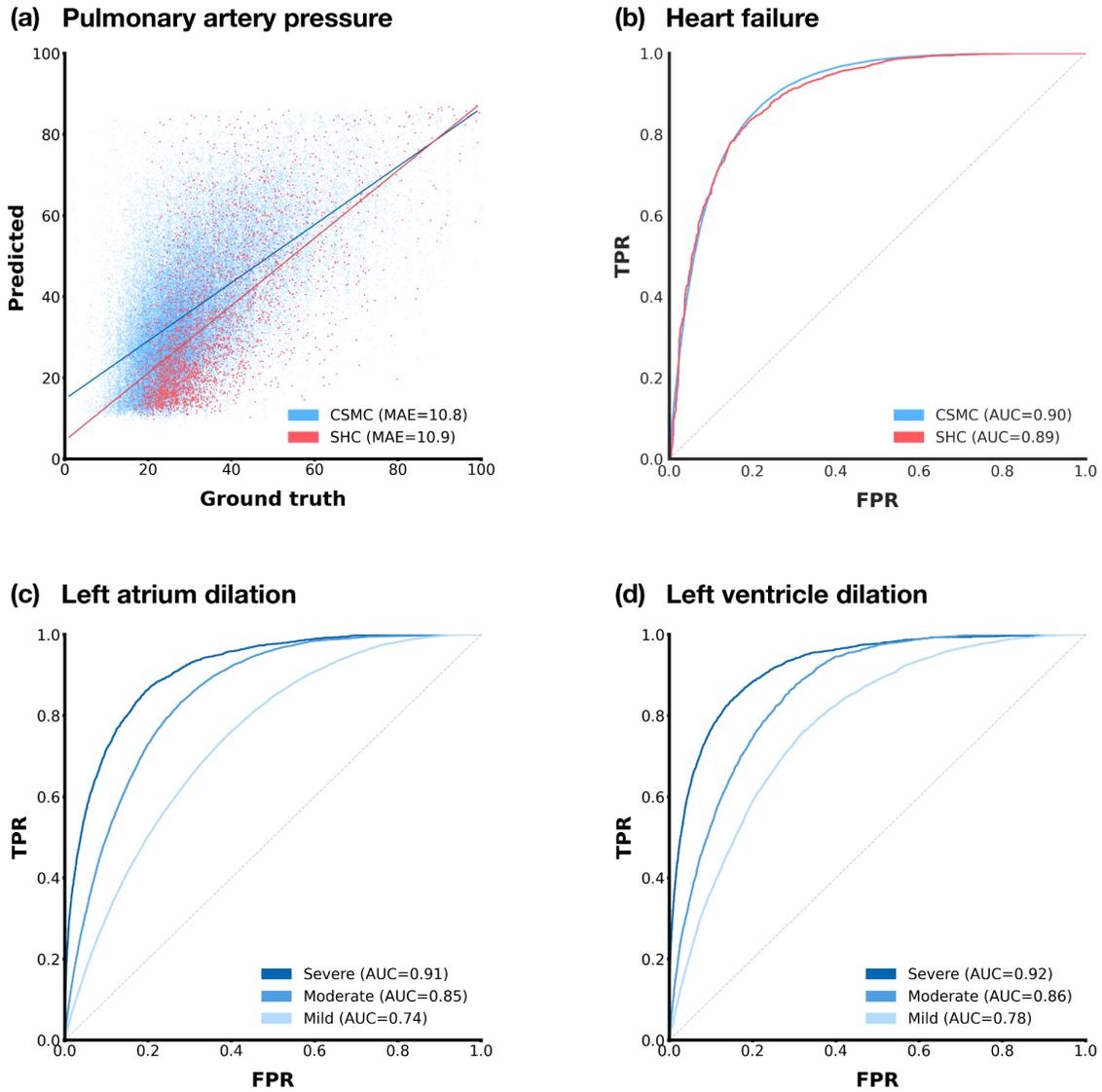

**Figure 3.** Zero shot model performance on held-out test apical-4-chamber videos from Cedars-Sinai Medical Center (blue, n = 100,994) and Stanford Healthcare (red, n = 5,000). (a) Estimation of pulmonary artery pressure. (b) Detection of heart failure with reduced ejection fraction. (c) Detection of left atrial dilation. (d) Detection of left ventricular dilation.

**Figure 4**

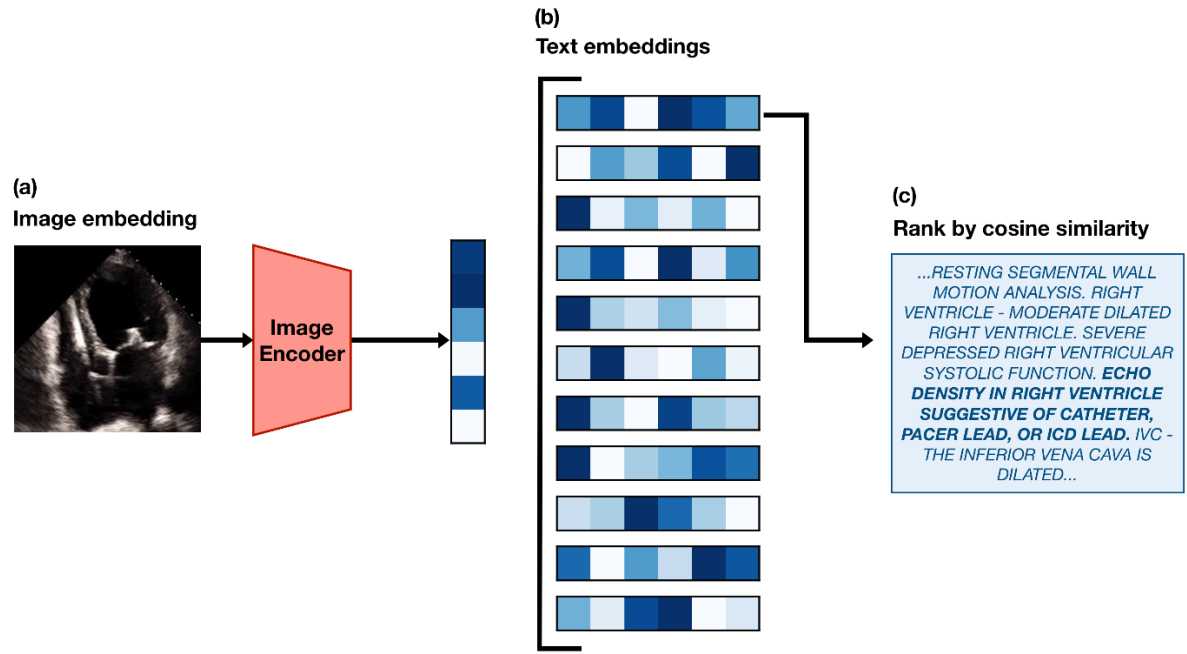

**Figure 4.** Image-to-text semantic search. (a) The query image is first embedded using EchoCLIP-R's image encoder. (b) Then, the similarities between this query embedding and the embeddings of all 21,484 unique text reports in the test set are computed. (c) The reports are ranked by their similarity to the query image embedding, and the report with the highest similarity is retrieved.

**Figure 5**

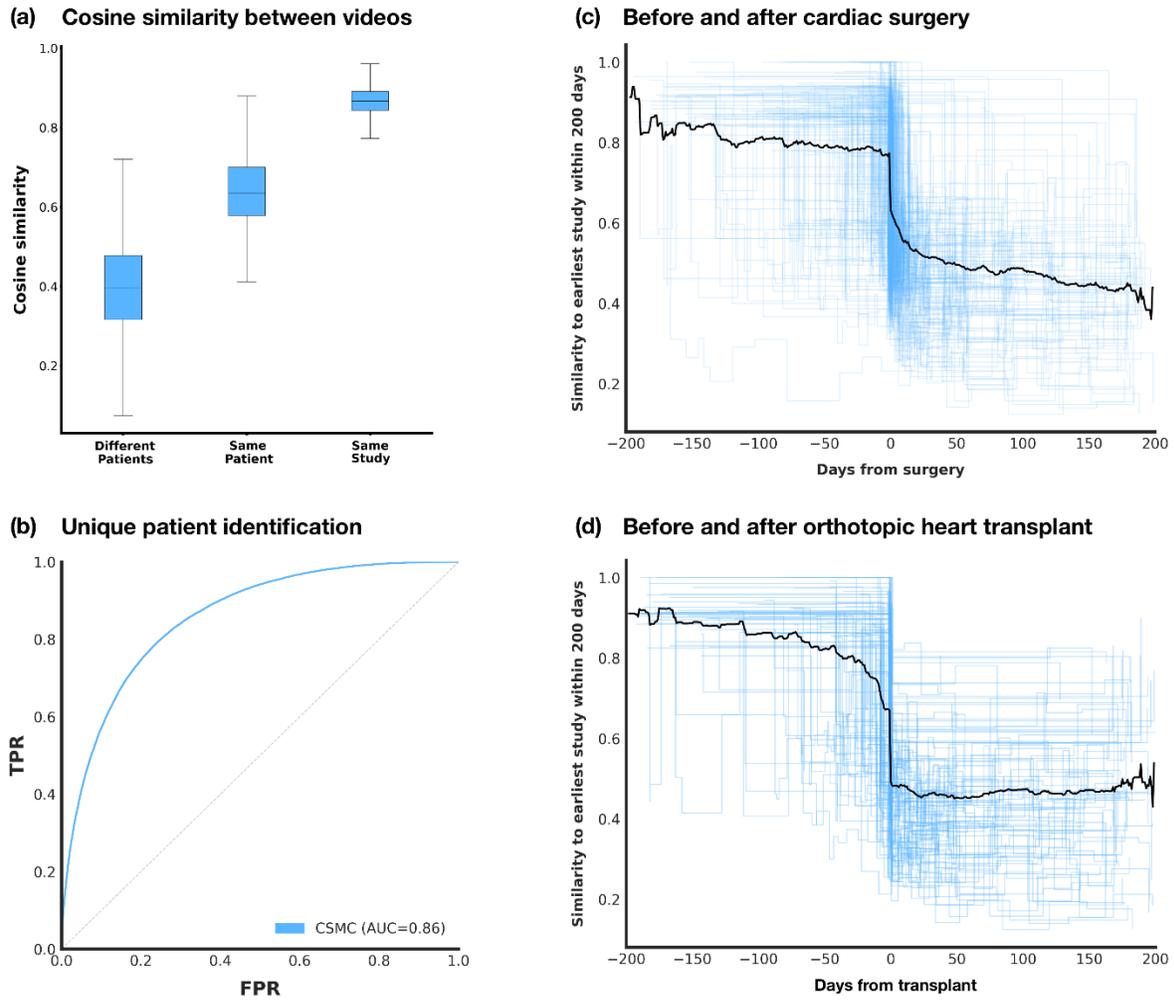

**Figure 5.** (a) Average similarity between embeddings from different patients is lower than between embeddings from the same patient. Average similarity between embeddings from the same patient is lower than between embeddings from the same study. Center lines indicate median, boxes span from the first to the third quartile, and whiskers stretch 1.5 times the inter-quartile range. (b) The similarity between image embeddings predicts whether the images come from the same patient with an AUC of 0.86 (95% CI: 0.86 - 0.86). (c,d) Plotting the similarity of videos acquired from the same patient over a window of 400 days to the first video acquired within the window. Patients either had a heart transplant (d) or other major cardiac surgery (c), and the 400-day window is centered on the date of these events.

**Supplementary Table 1.**

|  | Image encoder | Tokenization | Mean text-to-image rank | Mean image-to-text rank | Text-to-image retrieval, R@10 | Image-to-text retrieval, R@10 | MCMRR |
|---|---|---|---|---|---|---|---|
| *CLIP* | ViT-B-32 | CLIP BPE | 10,743.70 | 10,742.20 | 0.001 | 0 | 10,743.00 |
| *EchoCLIP* | ConvNeXt-Base | CLIP BPE | 560.4 | 582.1 | 0.16 | 0.166 | 571.3 |
| *EchoCLIP-R* | ConvNeXt-Base | Template tokenizer | **203.1** | **209.1** | **0.343** | **0.333** | **206.1** |
| *ViT backbone* | ViT-B-32 | CLIP BPE | 619.3 | 621.6 | 0.147 | 0.15 | 620.5 |
| *Random initialization* | ViT-B-32 | CLIP BPE | 867.4 | 863.7 | 0.106 | 0.112 | 865.6 |
| *With patch dropout* | ViT-B-32 | CLIP BPE | 829.0 | 833.5 | 0.11 | 0.118 | 831.2 |
| *EchoBPE* | ConvNeXt-Base | EchoBPE1K | 291.4 | 260.2 | 0.288 | 0.288 | 275.8 |

|  | Ejection fraction, MAE | Heart failure, AUC | Pacemaker, AUC | Impella, AUC | TAVR, AUC | MitraClip, AUC |
|---|---|---|---|---|---|---|
| *CLIP* | 20.76 (20.69-20.83) | 0.49 (0.49-0.50) | 0.51 (0.51-0.52) | 0.63 (0.60-0.67) | 0.46 (0.46-0.47) | 0.53 (0.52-0.54) |
| *EchoCLIP* | **8.35 (8.30-8.40)** | **0.90 (0.89-0.90)** | **0.84 (0.84-0.84)** | 0.98 (0.98-0.99) | **0.92 (0.91-0.92)** | **0.97 (0.97-0.97)** |
| *EchoCLIP-R* | 16.89 (16.83-16.95) | 0.50 (0.50-0.51) | 0.66 (0.65-0.66) | 0.42 (0.38-0.46) | 0.52 (0.51-0.52) | 0.81 (0.81-0.82) |
| *ViT backbone* | 9.61 (9.56-9.66) | 0.87 (0.87-0.87) | **0.84 (0.84-0.85)** | 0.99 (0.98-0.99) | 0.91 (0.91-0.92) | **0.97 (0.97-0.97)** |
| *Random initialization* | 11.83 (11.77-11.90) | 0.70 (0.69-0.70) | 0.69 (0.68-0.69) | 0.97 (0.96-0.97) | 0.83 (0.83-0.84) | 0.94 (0.94-0.95) |
| *With patch dropout* | 11.65 (11.58-11.71) | 0.77 (0.77-0.77) | 0.73 (0.72-0.73) | 0.91 (0.89-0.93) | 0.80 (0.80-0.81) | 0.89 (0.89-0.90) |
| *EchoBPE* | 10.28 (10.23-10.35) | 0.76 (0.76-0.76) | 0.77 (0.77-0.78) | 0.99 (0.98-0.99) | 0.86 (0.86-0.87) | 0.96 (0.96-0.96) |

|  | Significantly elevated RAP, AUC | Severe LV dilation, AUC | Severe RV dilation, AUC | Severe LA dilation, AUC | Severe RA dilation, AUC | PA pressure, MAE | Severely elevated PA pressure, AUC |
|---|---|---|---|---|---|---|---|
| *CLIP* | 0.53 (0.53-0.54) | 0.58 (0.57-0.59) | 0.56 (0.55-0.57) | 0.54 (0.53-0.55) | 0.59 (0.57-0.61) | 16.84 (16.77-16.91) | 0.52 (0.51-0.52) |
| *EchoCLIP* | **0.83 (0.82-0.83)** | **0.92 (0.92-0.93)** | **0.92 (0.91-0.92)** | 0.91 (0.90-0.92) | **0.97 (0.97-0.98)** | **10.81 (10.75-10.88)** | **0.85 (0.84-0.85)** |
| *EchoCLIP-R* | 0.48 (0.48-0.49) | 0.87 (0.86-0.87) | 0.81 (0.80-0.82) | 0.83 (0.82-0.84) | 0.81 (0.80-0.82) | 17.45 (17.38-17.52) | 0.61 (0.60-0.61) |
| *ViT backbone* | 0.82 (0.81-0.82) | 0.90 (0.90-0.91) | 0.91 (0.90-0.91) | **0.93 (0.92-0.93)** | 0.96 (0.96-0.97) | 11.26 (11.19-11.34) | 0.81 (0.81-0.81) |
| *Random initialization* | 0.66 (0.65-0.66) | 0.90 (0.89-0.91) | 0.82 (0.81-0.83) | 0.87 (0.86-0.87) | 0.90 (0.89-0.91) | 15.66 (15.58-15.73) | 0.63 (0.62-0.63) |
| *With patch dropout* | 0.63 (0.62-0.64) | 0.86 (0.86-0.87) | 0.77 (0.76-0.78) | 0.85 (0.84-0.86) | 0.89 (0.88-0.90) | 19.81 (19.73-19.89) | 0.59 (0.58-0.59) |
| *EchoBPE* | 0.69 (0.69-0.70) | 0.91 (0.90-0.91) | 0.84 (0.84-0.85) | 0.86 (0.85-0.86) | 0.93 (0.92-0.94) | 22.66 (22.58-22.74) | 0.80 (0.80-0.81) |